\documentclass[journal]{IEEEtran}

\ifCLASSINFOpdf
\else
\fi

\usepackage{epsfig}
\usepackage{algorithm}
\usepackage{algorithmic}
\usepackage{amsmath}
\usepackage{amssymb}
\usepackage{mathrsfs}
\usepackage{bm}
\usepackage{array,multirow}
\usepackage{mdwlist}
\usepackage{paralist}
\usepackage{url}
\usepackage{color}
\usepackage{caption}
\usepackage{setspace}
\usepackage{textcomp}
\usepackage{bbding}
\usepackage{booktabs}
\usepackage{threeparttable}
\usepackage{subfigure}
\usepackage[colorlinks=true,linkcolor=black,anchorcolor=black,citecolor=black]{hyperref}
\usepackage{arydshln}
\usepackage{animate}
\newcolumntype{L}[1]{>{\raggedright\arraybackslash}p{#1}}
\newcolumntype{C}[1]{>{\centering\arraybackslash}p{#1}}
\newcolumntype{R}[1]{>{\raggedleft\arraybackslash}p{#1}}

\usepackage{graphicx}
\graphicspath{{./figures/}}
\usepackage{cite}
\usepackage{verbatim}
\usepackage{makecell}

\usepackage{ulem}
\usepackage{color}
\definecolor{b}{RGB}{102,0,255}

\begin{document}
\title{Drone-based RGB-Infrared Cross-Modality Vehicle Detection via Uncertainty-Aware Learning}

\author{Yiming~Sun,
        Bing~Cao,
        Pengfei~Zhu,
        and~Qinghua~Hu,~\IEEEmembership{Senior Member,~IEEE}
\thanks{Yiming~Sun, Bing~Cao, Pengfei~Zhu and Qinghua~Hu are with the College of Intelligence and Computing, Tianjin University, Tianjin 300403, China (e-mail: sunyiming1895@tju.edu.cn; caobing@tju.edu.cn; zhupengfei@tju.edu.cn; huqinghua@tju.edu.cn). }
}


\maketitle

\begin{abstract}
Drone-based vehicle detection aims at finding the vehicle locations and categories in an aerial image. It empowers smart city traffic management and disaster rescue.
Researchers have made mount of efforts in this area and achieved considerable progress. Nevertheless, it is still a challenge when the objects are hard to distinguish, especially in low light conditions.
To tackle this problem, we construct a large-scale drone-based RGB-Infrared vehicle detection dataset, termed {\it DroneVehicle}. Our DroneVehicle collects $28,439$ RGB-Infrared image pairs, covering urban roads, residential areas, parking lots, and other scenarios from day to night.
Due to the great gap between RGB and infrared images, cross-modal images provide both effective information and redundant information.
To address this dilemma, we further propose an uncertainty-aware cross-modality vehicle detection (UA-CMDet) framework to extract complementary information from cross-modal images, which can significantly improve the detection performance in low light conditions.
An uncertainty-aware module (UAM) is designed to quantify the uncertainty weights of each modality, which is calculated by the cross-modal Intersection over Union (IoU) and the RGB illumination value.
Furthermore, we design an illumination-aware cross-modal non-maximum suppression algorithm to better integrate the modal-specific information in the inference phase.
Extensive experiments on the DroneVehicle dataset demonstrate the flexibility and effectiveness of the proposed method for cross-modality vehicle detection.
The dataset can be download from
\url{https://github.com/VisDrone/DroneVehicle}.

\end{abstract}


\IEEEpeerreviewmaketitle

\begin{figure}
    \centering
    \subfigure[Uncertainty of RGB modality.] {
        \label{fig:a}
        \includegraphics[width=0.45\columnwidth]{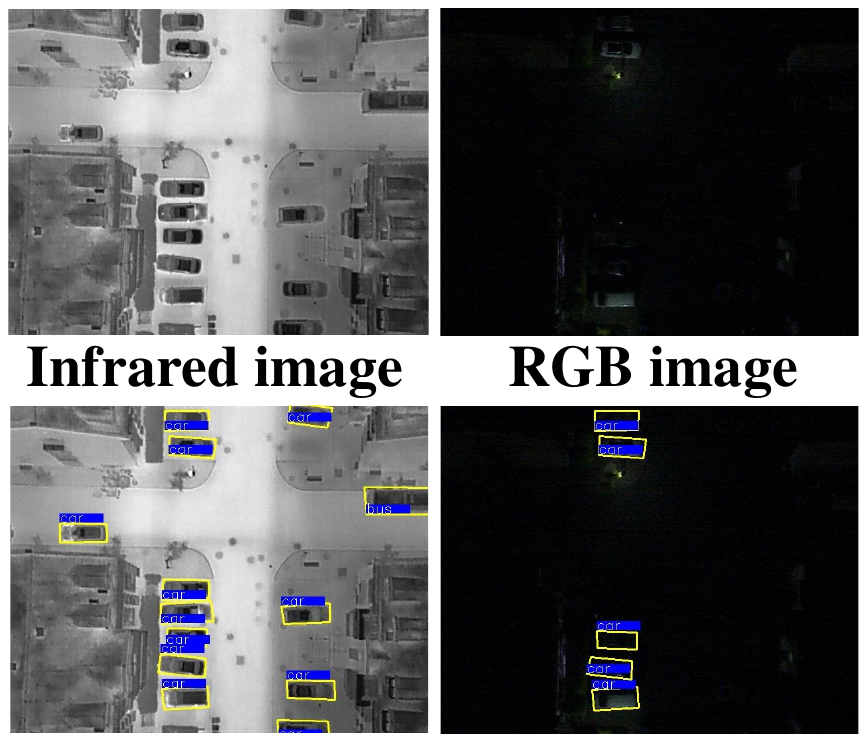}
    }
    \subfigure[Uncertainty of infrared modality.] {
        \label{fig:b}
        \includegraphics[width=0.45\columnwidth]{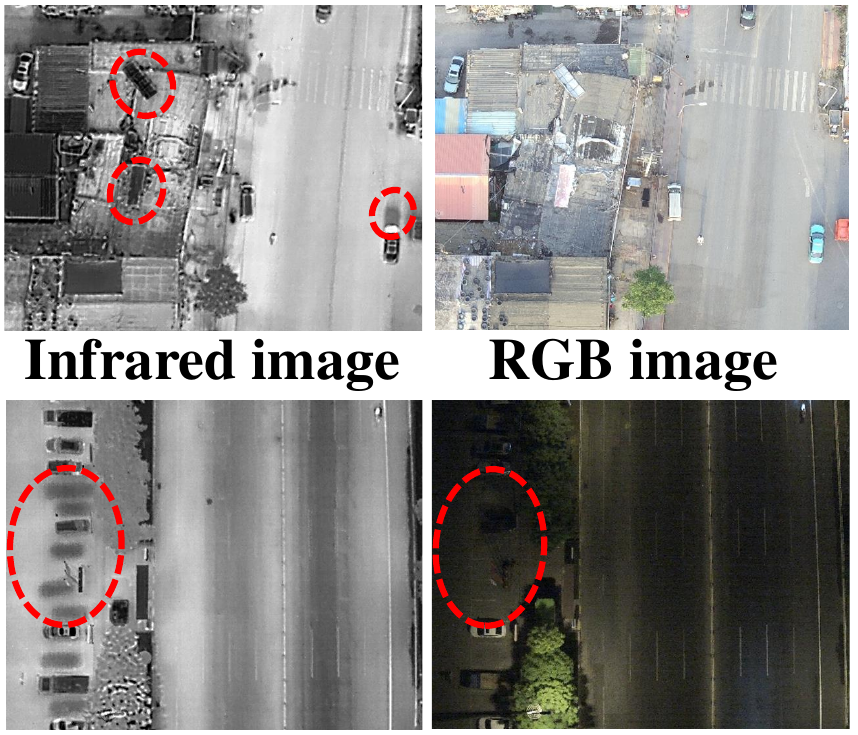}
    }
    \subfigure[Results in RGB modality.] {
        \label{fig:c}
        \includegraphics[width=0.45\columnwidth]{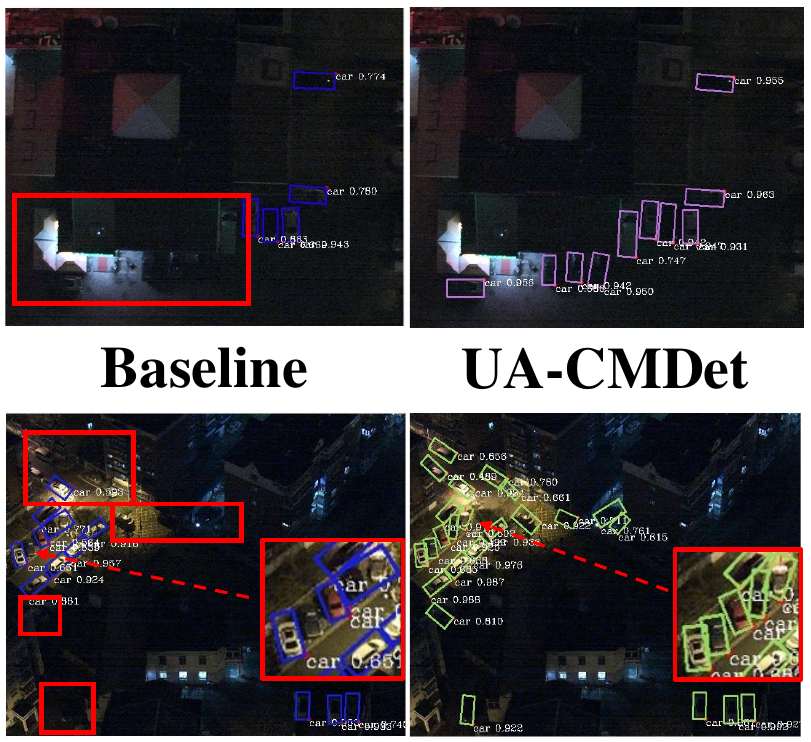}
    }
    \subfigure[Results in infrared modality.] {
        \label{fig:d}
        \includegraphics[width=0.45\columnwidth]{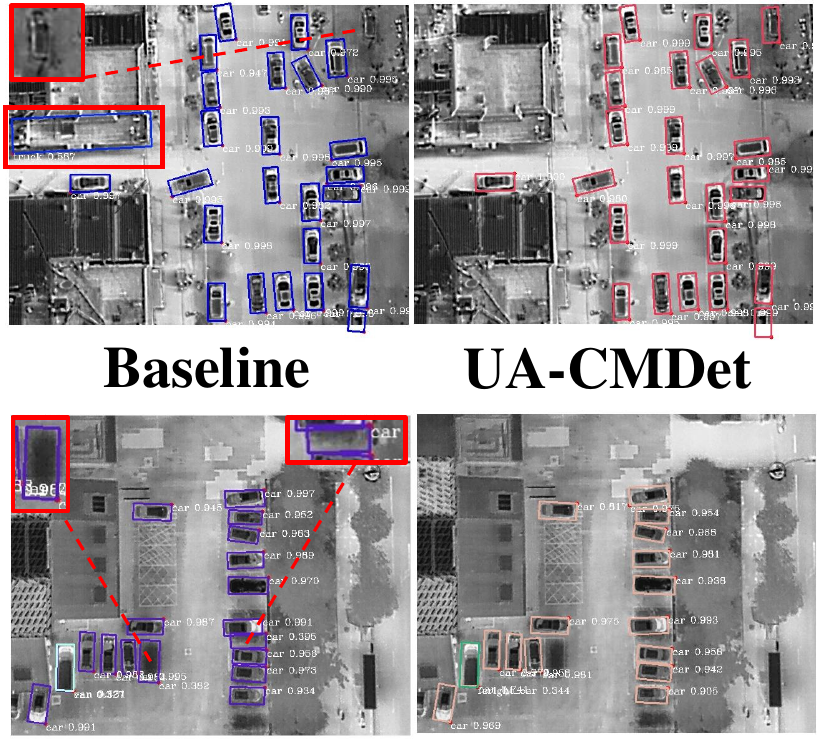}
    }
    \caption{ Uncertainty in the RGB modality and infrared modality. (a) At night, some RGB images that are completely out of light in aerial scenes, which leads to uncertain vehicle locations. In this situation, the corresponding infrared images can provide more clear imaging details. (b) Due to the lack of color information and the complex background of aerial scenes, there are some confusing rectangle objects in infrared images that look similar to vehicles. Besides, due to the thermal crossover of infrared imaging devices, ``\textit{ghost shadows}'' will appear in some locations without any vehicles. These problems cause strong uncertainty when we use infrared images in object detection, and affect the detection accuracy. In (c) and (d), the left side shows the results of the baseline on a single modality, and the right side shows the results of our method. Our method integrates the effective information of two modalities and achieves much better results.}
    \label{fig1}
\end{figure}

\section{Introduction}

\IEEEPARstart{D}{rone}-based vehicle detection plays an important role in smart city traffic management and disaster rescue~\cite{wen2021detection,zhu2020vision,8782625,9298794,8054712}.
Camera-equipped drones can collect images with a wider angle of view, which is more conducive to capturing objects on the ground.
However, due to the highly complex backgrounds and different illumination, object detection based on aerial images~\cite{Xia_2018_CVPR,7726065,zhu2018vision,6851187} is still an active and challenging task in computer vision.

Recently, some works~\cite{ding2019learning,yang2019scrdet,yang2020arbitrary,8651485,yang2021r3det,9001201} related to aerial images object detection has appeared.
These methods are designed for the RGB modality alone, which cannot cope with the challenges in low-light complex scenarios. But for smart city traffic management, disaster rescue, and other applications, a method that can deal with complex scenarios at all times is even more needed~\cite{8950077,8293689,9184996,8788689}.
At present, the biggest obstacle to this demand is the lack of large-scale full-time aerial view datasets for vehicle detection.
Considering the robustness of infrared cameras in full-time imaging, we attempt to introduce infrared images to provide complementary information for the RGB modality and form RGB-Infrared image pairs. To fill this gap, we collect a large-scale drone-based RGB-Infrared cross-modality vehicle detection dataset named {\it DroneVehicle}, which contains $28,439$ sets of RGB-Infrared image pairs, and provides oriented bounding box annotations for a total of $953,087$ objects.
In addition, our DroneVehicle covers multiple scenarios including urban roads, residential areas, parking lots, and other scenarios from day to night. To the best of our knowledge, this is the first and the largest full-time drone-based RGB-Infrared cross-modality dataset.

These RGB-Infrared cross-modality images, while introducing effective information, also introduce redundant information~\cite{9426573,9165167,9000872}.
For example, infrared images are not sensitive to light, which can provide more effective object locations and categories for RGB images under low-light conditions.
However, due to the lack of color information and the adverse effect of thermal crossover, some false objects with a similar appearance to real objects may appear in infrared images, which are redundant information for RGB images under good lighting conditions.
How to handle this dilemma and take the respective advantages of the RGB and infrared modalities to jointly improve the performance of object detection has attracted widespread attention from the community.

Some researchers have conducted in-depth research on RGB-Infrared cross-modality object detection~\cite{li2019illumination,zhang2019cross,2020Task,zhou2020improving,9419080}.
Unfortunately, these methods are only designed for multispectral pedestrian detection tasks, which cannot predict the orientations of the objects in aerial scenes. Therefore, these methods cannot handle the detection of vehicles with different orientations and different categories in aerial scenes. In addition, these works are usually carried out on urban street view scenes, which are unable to address certain problems in aerial scenes.
In addition to the different orientations of objects in aerial scenes mentioned above, there are also problems such as wide coverage scenes, confusing image background, and extremely low visibility of some scenarios.

\begin{table*}
  \renewcommand\arraystretch{0.9}
      \caption{Comparison of the state-of-the-art benchmarks and datasets. Note that, the resolution indicates the maximum resolution of videos/images included in the benchmarks or datasets, R stands for RGB modality, I stands for infrared modality, and the BB is short for bounding box. ($1k=1,000$)}
      \begin{center}
      \begin{tabular}{C{3.3cm}C{1.4cm}C{1.15cm}C{1.1cm}C{1.25cm}C{2.55cm}C{1.9cm}C{1.1cm}C{0.8cm}}
      \toprule
        {Object detection datasets}  &\multirow{2}{*}{Scenario}   &\multirow{2}{*}{Modality}     &\multirow{2}{*}{\#Images}    &\multirow{2}{*}{Categories}        &Avg. \#labels/categories     &\multirow{2}{*}{Resolution}      &{Oriented BB}    &\multirow{2}{*}{Year} \\
      \hline
      \specialrule{0em}{1pt}{1pt}
		KITTI~\cite{Geiger2012Are}   &driving  &R  &$15.4k$  &$2$  &$80k$  &$1241\times376$  &  &2012 \\
		PASCAL VOC~\cite{EveringhamThePascalVisual}   &life  &R  &$16.5k$  &$20$ &$2,002$  &$469\times387$  &   &2012 \\
		MS COCO~\cite{LinMicrosoft}   &life &R  &$328.0k$  &$80$  &$31.2k$  &$640\times640$  &   &2014 \\
		DLR~3K~\cite{liu2015fast}    &aerial  &R  &$20$   &$2$  &$2,946$   &$5616\times3744$   &$\surd$    &2015 \\
		VEDAI~\cite{RazakarivonyVehicle}    &aerial  &R/I  &$1.2k$   &$9$   &$411$ &$1024\times1024$ &$\surd$    &2015 \\
		UA-DETRAC~\cite{Wen2015UA}    &surveillance  &R  &$140.1k$  &$4$  &$302.5k$  &$960\times540$ &    &2015 \\
		COWC~\cite{Mundhenk2016A}   &aerial  &R  &$32.7k$  &$1$ &$32.7k$ &$2048\times2048$ &   &2016\\
		CARPK~\cite{Hsieh2017Drone}    &drone  &R  &$1,448$  &$1$  &$89.8k$  &$1280\times720$  &    &2017 \\
		DOTA~\cite{Xia_2018_CVPR}    &aerial  &R  &$2,806$  &$15$  &$12.5k$  &$12029\times5014$   &$\surd$  &2018 \\
		UAVDT~\cite{du2018unmanned}    &drone  &R  &$80k$  &$3$  &$280.5k$  &$1080\times540$  &   &2018 \\
		VisDrone~\cite{zhu2018vision}    &drone  &R  &$10,209$  &$10$  &$54.2k$  &$2000\times1500$  &   &2018 \\
		BDD100K~\cite{yu2020bdd100k}   &driving  &R  &$100k$  &$10$   &$184k$  &$1280\times720$ &  &2020 \\	EAGLE~\cite{azimi2021eagle}    &aerial  &R  &$8,280$  &$2$  &$107.9k$  &$936\times936$   &$\surd$  &2020 \\
      \hdashline\specialrule{0em}{1pt}{1pt}
		\textbf{DroneVehicle(ours)}    &drone  &R+I   &$56,878$  &$5$  &$190.6k$  &$840\times712$  &$\surd$   &2021 \\
      \toprule
      \end{tabular}
      \end{center}
      \label{tab:comparison-detection dataset}
\end{table*}

Our DroneVehicle shows that the RGB-Infrared cross-modality vehicle detection still faces great challenges in aerial scenes.
Fig.~\ref{fig1}~\subref{fig:a} shows some RGB images that are completely out of light in aerial scenes, which leads to uncertain vehicle locations. In this situation, the corresponding infrared images can provide more clear imaging details. Nevertheless, due to the color information lacking, the infrared images also perform poor detection accuracy in some daytime scenarios with good illumination. For instance, some confusing rectangle objects in the infrared image look similar to vehicles, as shown in the first row of Fig.~\ref{fig1}~\subref{fig:b}.
Besides, due to the thermal crossover of infrared imaging devices, ``\textit{ghost shadows}'' will appear in some locations without any vehicles. These problems cause strong uncertainty when we use infrared images in object detection, and affect the detection accuracy.
Moreover, due to the subtle difference between the poses of the visible light camera and the infrared camera during image collection, there may be some pixel misalignment between the RGB and infrared images, which also brings cross-modality uncertainty in object position.

\begin{figure}[!t]	
	\centering	
	\includegraphics[width=0.98\columnwidth]{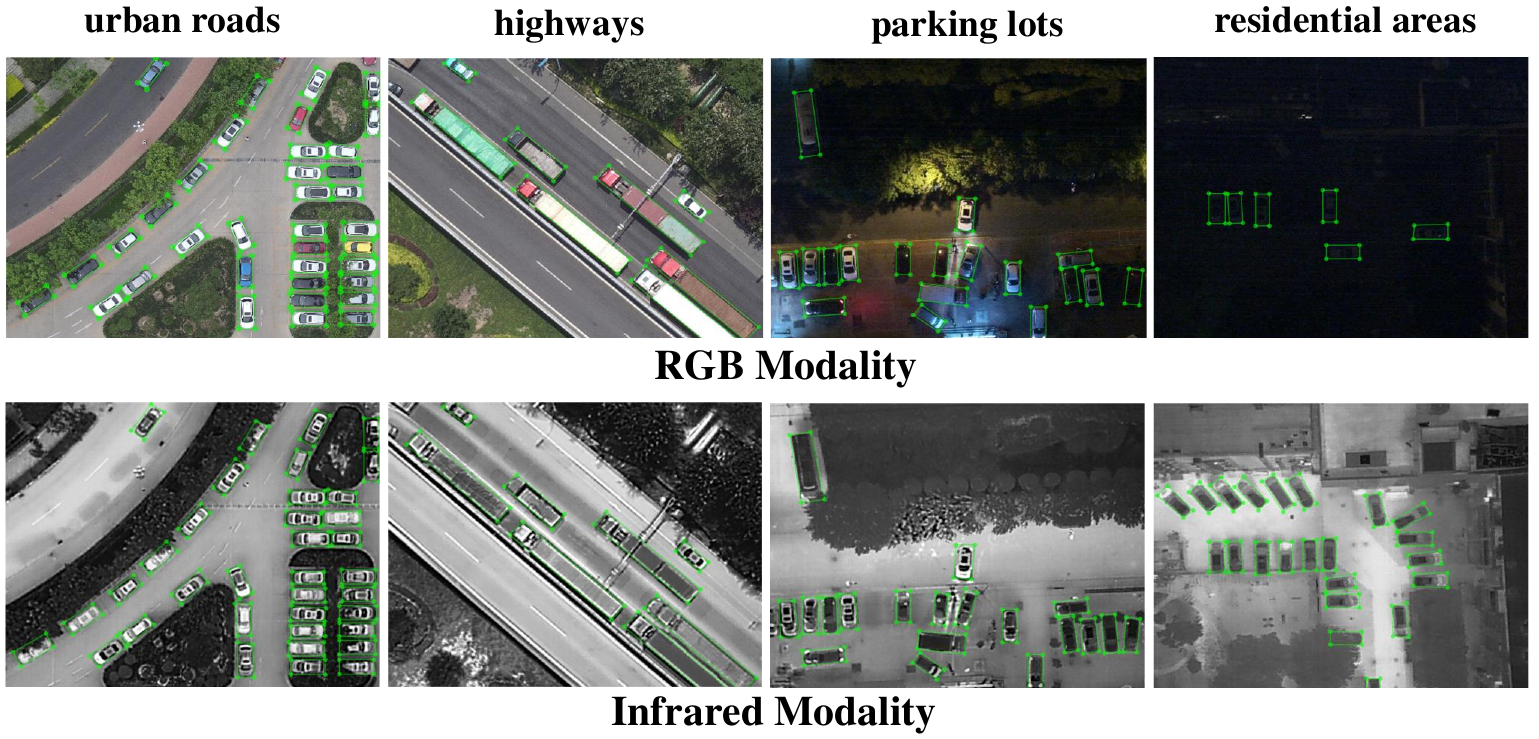}
	\caption{Some example annotated images of the DroneVehicle dataset. The first row shows some examples in the RGB modality, and the second row shows the corresponding examples in the infrared modality. }
	\label{fig6}
\end{figure}

To tackle this problem, we propose an uncertainty-aware cross-modality vehicle detection (UA-CMDet) method, which combines RGB and Infrared information in a unified framework.
Specifically, we design an uncertainty-aware module (UAM) to quantify the uncertainty of each modality. The UAM takes the ground-truth annotation of RGB-Infrared image pairs as prior knowledge and joins the RGB illumination and the cross-modality Intersection over Union (IoU) to calculate the uncertainty weights of each modality. Note that, this module is removed after training, and it does not increase any extra computations during inference.
In addition, we further design an illumination-aware cross-modal non-maximum suppression (IA-NMS) strategy to fuse the detection results of different modalities.

The effectiveness and reliability of the proposed methods are verified on our DroneVehicle dataset. As shown in Fig.~\ref{fig1}~\subref{fig:c}, in the RGB modality, due to poor lighting conditions, many objects are not detected by the baseline, and multiple objects are even mistakenly detected as one.
In addition, due to the lack of color and texture details, the infrared images appear ``ghost shadows'' caused by thermal crossover, which is hard to detect correctly by the baseline methods. As a comparison, our method integrates RGB information to infrared modality and achieves much better results, as shown in Fig.~\ref{fig1}~\subref{fig:d}.

The main contributions of this paper are summarized as follows:
\begin{itemize}
\item We construct a large-scale drone-based RGB-Infrared dataset ({\it DroneVehicle}) that contains $953,087$ object instances in $56,878$ images recorded from the diverse scenarios. Compared to other drone-based datasets, our DroneVehicle is the first and the largest dataset that can significantly promote the development of drone-based cross-modal vehicle detection.
\item We propose an uncertainty quantification method, termed uncertainty-aware module (UAM), to effectively measure the uncertainty between two modalities by data prior information. Our UAM can be flexibly applied to various cross-modality object detection frameworks.
\item We propose an uncertainty-aware cross-modality detector (UA-CMDet), which for the first time joint the uncertainty information of two different modalities to boost the vehicle detection performance and achieve superior performance against state-of-the-arts.
\end{itemize}

The remaining paper is organized as follows. We briefly summarize the related works in Sec.~\ref{sec:2}. Our \textit{DroneVehicle} dataset is introduced in Sec.~\ref{sec:3}. We present the proposed framework (UA-CMDet) in Sec.~\ref{sec:4} and conducted extensive experiments in Sec.~\ref{sec:5} to validate its effectiveness. Sec.~\ref{sec:6} concludes this work.

\section{Related Work}
\label{sec:2}
In this section, we first review several natural scene vehicle detection datasets, then briefly review the existing aerial view datasets that can support vehicle detection tasks, and finally review some state-of-the-art vehicle detection algorithms.

\subsection{Existing Natural Scene Datasets}
In recent years, researchers have established many excellent natural scene datasets for evaluating vehicle detection. Specifically, it mainly includes the life scene datasets represented by PASCAL VOC~\cite{EveringhamThePascalVisual} and MS COCO~\cite{LinMicrosoft}, and the driving scene datasets represented by KITTI~\cite{Geiger2012Are} and BDD100K~\cite{yu2020bdd100k}.
PASCAL VOC is a well-known benchmark for object detection, which contains $16,551$ images and annotation files. It provides bounding box annotations for $20$ different categories. Among them, the categories with vehicle attribute are \textit{Bus} and \textit{Car}. According to~\cite{EveringhamThePascalVisual}, there are $3,267$ annotated bounding boxes for \textit{car} and $822$ annotated bounding boxes for \textit{Bus}.
MS COCO is a large dataset, which is mainly used for object detection and semantic segmentation. Compared with the PASCAL VOC in object detection, MS COCO has a more complex background, a larger number of objects, and a smaller size of objects. Therefore, the MS COCO is more challenging. MS COCO contains more than $328,000$ images with $2.5$ million manually annotated object instances. It has $80$ object categories with $31,200$ instances on average per category. But it only contains three categories with vehicle attributes (\textit{Car}, \textit{Bus} and \textit{Truck}), so the corresponding annotated bounding boxes are also limited.
KITTI and BDD100K are well-known driving scene datasets, which are designed to evaluate environmental perception algorithms in autonomous vehicles. They are all collected by vehicle cameras in an urban environment.
KITTI consists of $7,481$ training images and $7,518$ test images, and contains more than $200,000$ object annotations captured in cluttered scenarios. It mainly includes vehicle attribute annotations and pedestrian attribute annotations, among which vehicle attribute annotations specifically include \textit{cars}, \textit{vans}, \textit{trucks}, \textit{trams}.
BDD100K provides a total of $1,841,435$ annotated bounding boxes for $100,000$ key frames captured from $100,000$ videos. It has annotated 10 categories, of which there are three categories (\textit{Bus}, \textit{Truck}, \textit{Car}) with vehicle attribute. Although these datasets promote the advancement of vehicle detection technology in natural scenes, they cannot meet the needs of vehicle detection in scenarios such as smart city and disaster rescue. These scenes often need to deal with a broader perspective, more variable object orientations, and scales, so we propose an aerial view dataset specifically designed for vehicle detection to meet these needs.

\subsection{Existing Aerial View Datasets}
In recent years, some aerial view datasets have been proposed for aerial vehicle detection. UA-DETRAC\cite{Wen2015UA} is a large-scale dataset for vehicle detection and tracking. It is mainly shot on road crossing bridges in Beijing and Tianjin, China. It has more than $140,000$ frames and a total of $1.21$ million labeled bounding boxes of objects, of which $8250$ vehicles are manually annotated.
DLR~3K~\cite{liu2015fast} is composed of $20$ aerial images captured by the \textit{DLR~3K} camera system of the German Aerospace Center. It mainly contains two categories (\textit{Car} and \textit{Truck}), and each object is manually annotated as an oriented bounding box. Since the image was taken at a height of 1000 meters above the ground, the image resolution is set to $5616\times3744$ pixels.
VEDAI~\cite{RazakarivonyVehicle} is a database for evaluating the detection of small vehicles in aerial images. It includes different vehicle categories for a total of more than $3,700$ annotated objects in more than $1,200$ images and includes four different sub-sets (large-size color images, small-size color images, large-size infrared images, small-size infrared images).
COWC~\cite{Mundhenk2016A} is a large diverse dataset of cars from overhead images. It includes $32,716$ \textit{unique cars} and $58,247$ usable negative examples.
CARPK~\cite{Hsieh2017Drone} consists of $1,448$ images and contains $89,777$ annotated cars captured by the drone from different parking lots.
UAVDT\cite{du2018unmanned} mainly contains about $80,000$ representative frames from $10$ hours raw videos and annotated $14$ kinds of attributes ($e.g.$, weather condition, flying altitude, vehicle category, and occlusion) with bounding boxes.
VisDrone~\cite{zhu2018vision} consists of $263$ video clips and $10,209$ images with rich annotations (such as object bounding boxes, object categories, occlusion, truncation ratios, $etc.$). DOTA~\cite{Xia_2018_CVPR} includes $15$ different categories and contains $188,282$ annotated object instances. The latest DOTA-v1.5 also adds many annotations to the small object instances about or below $10$ pixels that were missed previously.
EAGLE~\cite{azimi2021eagle} is a large-scale dataset for multi-class vehicle detection with object orientation information in aerial imagery. It provides a total of $215,986$ bounding box annotations for the two main categories (small vehicles and large vehicles), including $208,963$ small and $7,023$ large vehicles.
The above-mentioned datasets have greatly promoted the application of vehicle detection in aerial view scenes, but there are many dark night scenarios in the real world, and it is difficult to handle such cases only relying on the RGB modality. However, all of the above-mentioned datasets only contain RGB modality except the VEDAI. The infrared images generally perform well in dark night scenarios, but unfortunately, the VEDAI dataset is small and lacks sufficient dark night scenarios, so the complementarity of the two modalities cannot be reflected.
In contrast to the above-mentioned aerial view datasets, our proposed DroneVehicle dataset contains pairs of RGB images and infrared images collected in various urban environments, which can support the research of cross-modal vehicle detection.

\begin{figure}[!t]
	\centering	
	\includegraphics[width=3.3in]{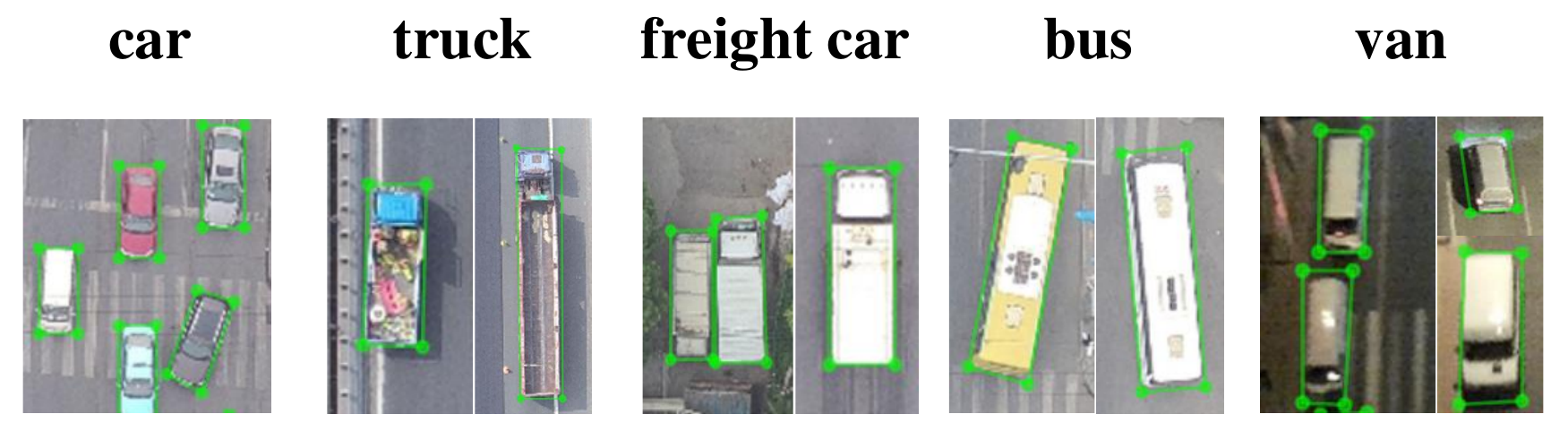}
	\caption{Examples of five categories in the DroneVehicle.}
	\label{fig13}
\end{figure}

\subsection{Vehicle Detection}
Vehicle detection aims to find the location of the vehicle and determine its category~\cite{dong2015vehicle}. In recent years, significant progress has been made in object detection. As an important topic in object detection, vehicle detection also benefits from the progress of object detection. Among them, whether it is natural scenes or aerial scenes, vehicle detection algorithms represented by RetinaNet~\cite{lin2017focal} and Faster R-CNN~\cite{ren2015faster} have been widely used.
RetinaNet is a dense detector consisting of the feature extractor, the classification subnetworks, and the regression subnetworks. Its main contribution is the proposed focal loss, which aims to solve the problem of the extreme imbalance between the foreground category and the background category of the dense detector during the training process.
Different from RetinaNet, Faster R-CNN is a two-stage detector. In the first stage, region proposals are first generated through RPN (Region Proposal Networks), and then these candidate regions are classified to distinguish foreground and background. And perform bounding box regression on these candidate regions. In the second stage, the feature maps of the region of interest are extracted through the RoI Pooling layer, and then sent to the classification subnetworks of the vehicle category and the regression subnetworks of the bounding box.
In aerial scenes, the orientation and scale of many vehicle objects change very frequently, so vehicle detectors based on Deformable RoI Pooling~\cite{dai2017deformable} can often achieve better performance. Compared with RoI Pooling in Faster R-CNN, the main difference of Deformable RoI Pooling is the addition of an offset learning module, so that the sampling points of the convolution kernel on the feature map can be offset to focus on the region of interest.
Mask R-CNN~\cite{he2017mask} adds a branch to predict the segmentation mask based on Faster R-CNN, and at the same time replaces the RoI Pooling layer with the RoI Align layer. Among them, the RoI Align layer alleviates the problem of misalignment between the feature map and the original image in the RoI feature extraction stage.
Cascade Mask R-CNN~\cite{cai2018cascade} is composed of multiple detectors, which are trained in stages with increasing IoU thresholds. A detector outputs a good data distribution as input and then continues to train the next detector. This method effectively improves the false positive problem.
Hybrid Task Cascade~\cite{chen2019hybrid} improves the information flow by incorporating cascade and multi-tasking at each stage and leverage spatial context to further boost the accuracy.
RoITransformer~\cite{ding2019learning} adds an RoI Transformer module based on Faster R-CNN. The module is mainly composed of RRoI Leaner and RRoI Wrapping. Its core idea is to convert the horizontal proposals HRoI output by RPN into the oriented proposals RRoI.
Although the above methods have promoted the progress of vehicle detection, they only support a single modality, especially in dark night scenarios, these algorithms will fail in the RGB modality. To tackle this problem, we propose an uncertainty-aware cross-modality vehicle detection (UA-CMDet) method, which combines RGB and Infrared information in a unified framework.

\begin{table}
  \renewcommand\arraystretch{0.9}
      \caption{The total number of annotated bounding boxes for each category in the two modalities.}
      \begin{center}
      \begin{tabular}{C{1.2cm}C{1.15cm}C{0.85cm}C{0.85cm}C{0.8cm}C{1.48cm}}
      \toprule
        Modality      &car     &truck    &bus     &van   &freight car\\
      \hline
      \specialrule{0em}{1pt}{1pt}
		RGB   &$389,779$     &$22,123$   &$15,333$    &$11,935$     &$13,400$\\
      \hdashline\specialrule{0em}{1pt}{1pt}
		Infrared   &$428,086$     &$25,960$    &$16,590$    &$12,708$     &$17,173$\\
      \specialrule{0em}{1pt}{1pt}
      \toprule
      \end{tabular}
      \end{center}
      \label{tab:annotated bounding boxes}
\end{table}

\section{DroneVehicle Dataset}
\label{sec:3}
We select five vehicle categories of frequent interests in drone applications, i.e., \textit{car}, \textit{truck}, \textit{bus}, \textit{van} and \textit{freight car}. We carefully annotated $953,087$ oriented bounding boxes of object instances from these categories. The detailed comparison of the provided drone datasets with other related benchmark datasets in object detection are presented in Table.~\ref{tab:comparison-detection dataset}.

\subsection{Data Collection}
The DroneVehicle dataset contains $28,439$ RGB-Infrared image pairs. All the images are captured by the camera-equipped drones, covering a wide range of aspects, including scenarios (different types of urban roads, residential areas, parking lots, highways, $etc.$) and objects (\textit{car}, \textit{bus}, \textit{truck}, \textit{van}, \textit{freight car}, $etc.$). Note that, the dataset was collected using drone platforms in different scenarios and under various lighting conditions. The DroneVehicle contains $953,087$ manually annotated bounding boxes. Some example images are shown in Fig.\ref{fig6}.

\subsection{Data Preprocessing}
\subsubsection{Data Pruning}
Data pruning is an important step in making a dataset, so the raw data collected by the drone needs to be pre-processed. A part of the images with poor imaging quality are discarded, such as blurred images. Then we manually check all the image data and uniformly convert the resolution of the image to $840\times712$. Finally, we get the data without annotation.

\subsubsection{Data Calibration}
We first perform distortion correction on all cleaned images.
Since the attitudes of the drone are difficult to maintain absolute stability during the data collection process, the cross-modal image pair captured by the two cameras will inevitably appear pixel misalignment. In the calibration stage, we perform affine transformation and region cropping on each RGB-Infrared image pair to ensure that most cross-modal image pairs are aligned.

\begin{figure}[!t]	
	\centering	
	\includegraphics[width=1\columnwidth]{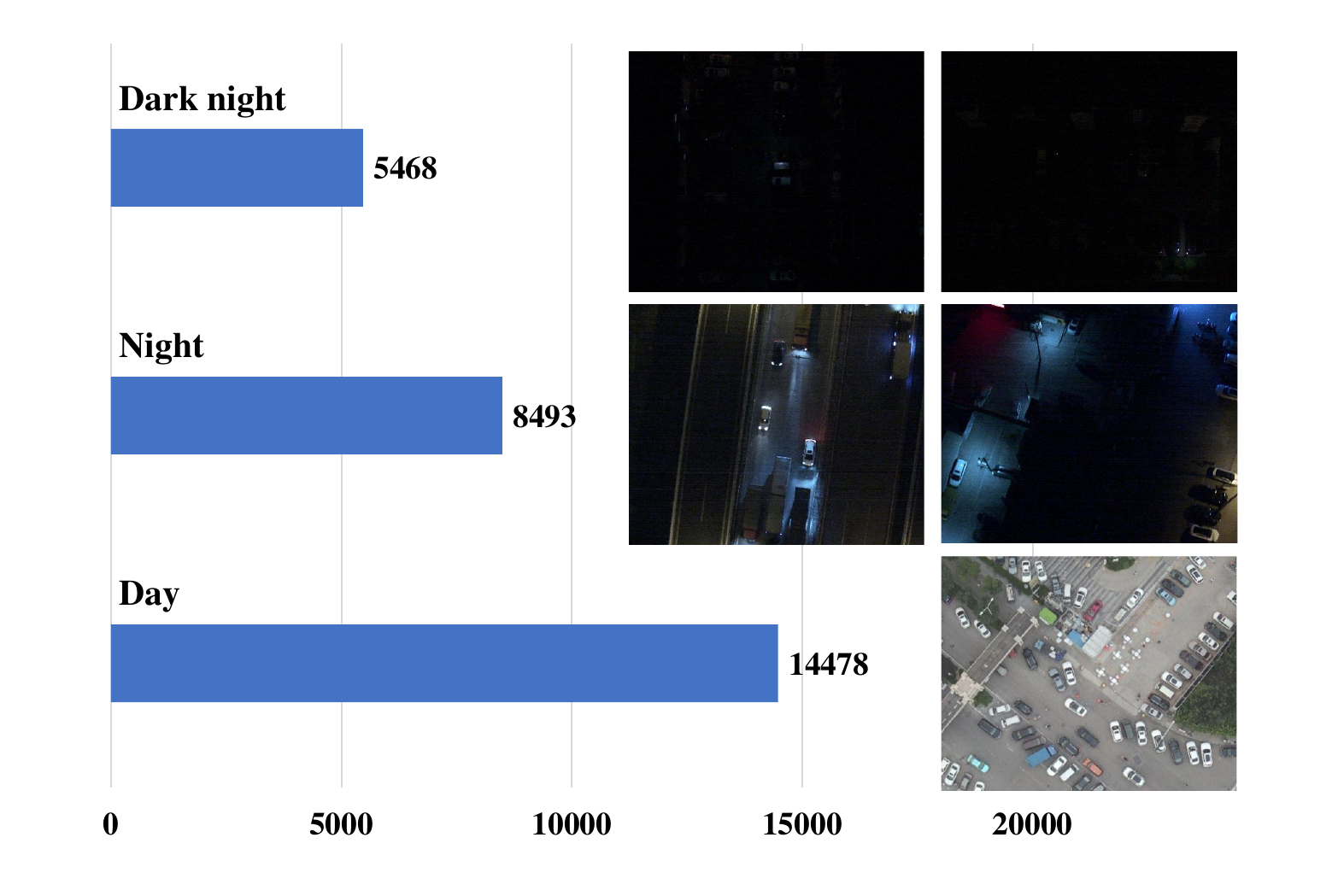}
	\caption{The distribution of data under different lighting conditions in the DroneVehicle dataset and its sample images.}
	\label{fig9}
\end{figure}

\begin{figure}[!t]
	\centering	
	\includegraphics[width=1\columnwidth]{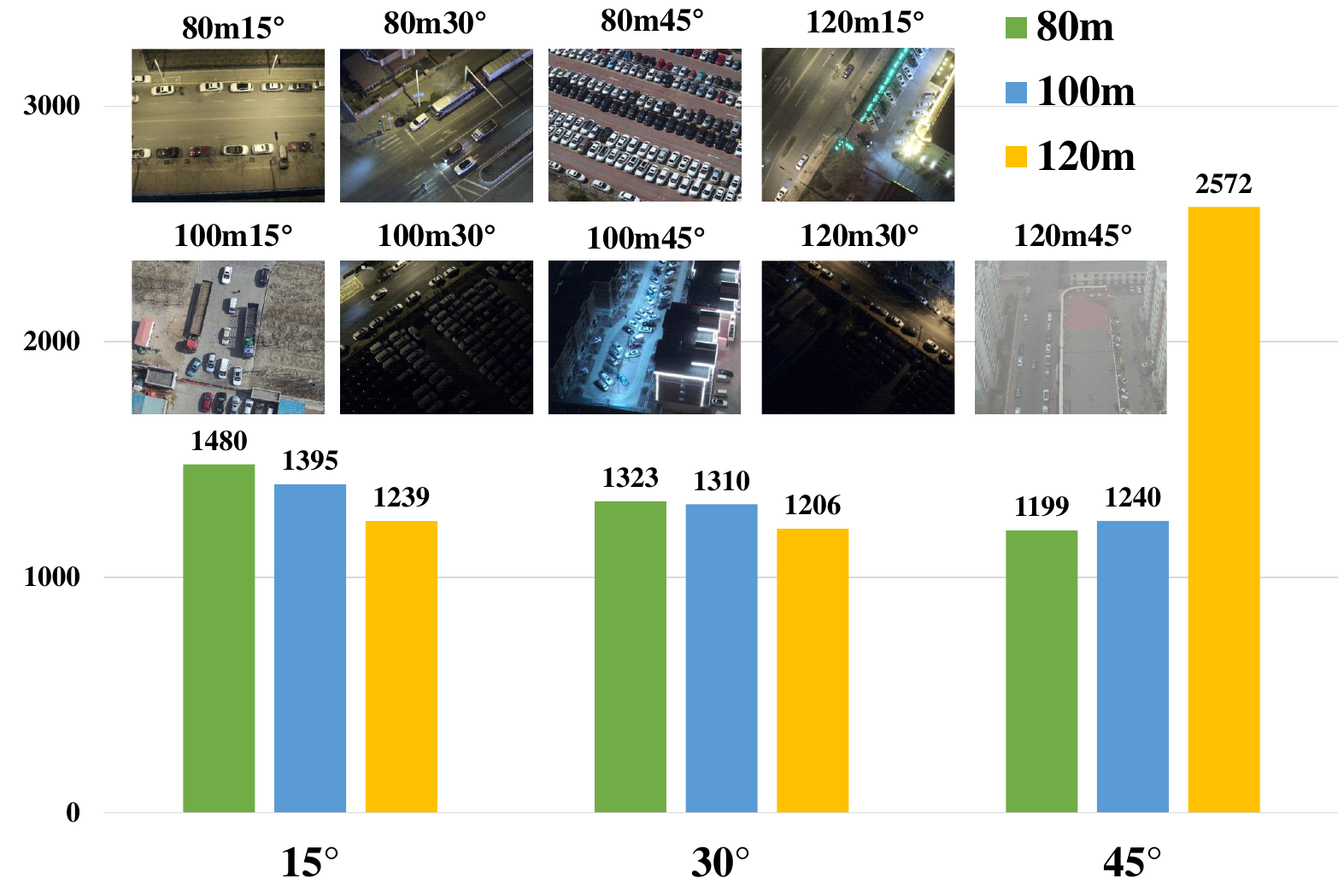}
	\caption{The distribution of data at different heights and angles in the DroneVehicle dataset and its sample images.}
	\label{fig12}
\end{figure}

\subsection{Data Annotation}
In computer vision, the typical annotation methods mainly utilize rectangular bounding boxes to annotate the objects on images. The bounding boxes are annotated with $(x_{c},y_{c},w,h)$, where $(x_{c},y_{c})$ is the center location, $w$ and $h$ are the width and height. These annotation methods are qualified for many scenarios, such as autonomous driving scenarios and traffic surveillance scenarios.
Due to the uncertainty of the camera angle in the UAV, it is difficult to calibrate the object precisely with these methods. Considering the various object orientations in aerial images, we deploy the oriented bounding box to accurately and compactly represent the object outline in the annotation procedure. Following \cite{Xia_2018_CVPR}, we choose the arbitrary quadrilateral bounding boxes to annotate oriented objects. In detail, we annotate $\{(x_{i},y_{i}),i = 1,2,3,4\}$ for each object, where $(x_{i},y_{i})$ denotes the vertices positions of the oriented bounding boxes in the image. Some samples of annotated images in our dataset are shown in Fig.\ref{fig6}.

\begin{figure*}[!t]
	\centering
	\includegraphics[width=2\columnwidth]{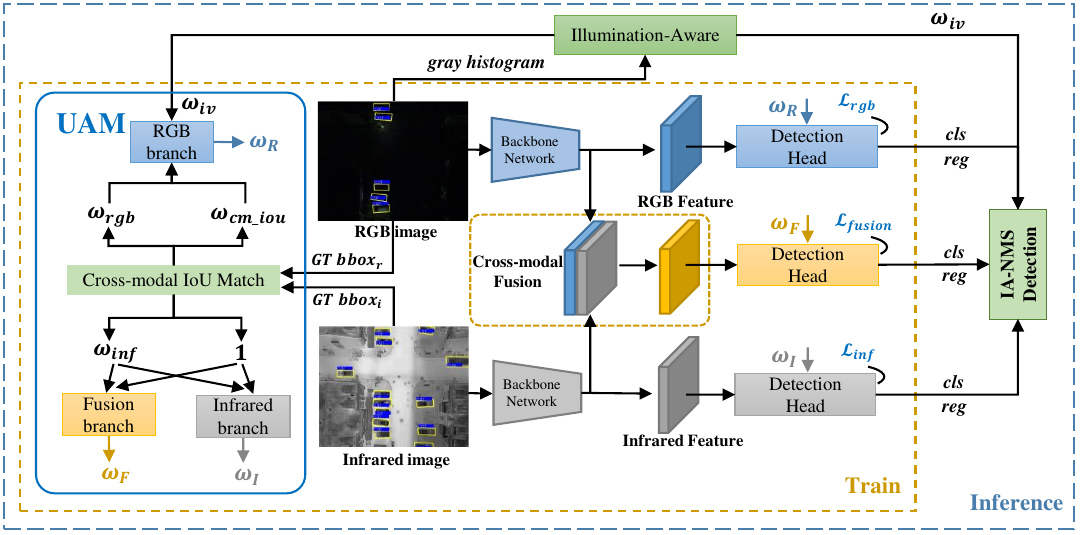}
	\caption{The architecture of the proposed UA-CMDet. UA-CMDet contains three branches. Among them, two branches take RGB image and infrared image respectively as input and the other branch takes the fused feature map as input. In the training phase, the three branches respectively predict their classification confidence scores and bounding box coordinates. Among them, in each detection head, the uncertainty weight of the corresponding modality is obtained through the uncertainty-aware module (UAM). And in the inference phase, the output of each branch will be post-processed through IA-NMS to obtain the final detection results. In UAM, first calculate the IoU between the ground-truth bounding boxes of the RGB image and the Infrared image. According to the cross-modal IoU matching results, UAM calculates three uncertainty weights for three different branches and outputs these uncertainty weights to the corresponding detection heads. }
	\label{fig2}
\end{figure*}

\subsection{Statistics and Attributes}
Five categories are chosen and annotated in our DroneVehicle dataset, including car, truck, bus, van, and freight car. Some examples are shown in Fig.\ref{fig13}. In the DroneVehicle dataset, we separately counted the total number of annotated bounding boxes for each category in RGB and infrared modalities. The statistical results are shown in Table.~\ref{tab:annotated bounding boxes}.

The DroneVehicle dataset consists of a total of $56,878$ images collected by the drone, half of which are RGB images, and the resting are infrared images. We have made rich annotations with oriented bounding boxes for the five categories. Among them, \textit{car} has $389,779$ annotations in RGB images, and $428,086$ annotations in infrared images, \textit{truck} has $22,123$ annotations in RGB images, and $25,960$ annotations in infrared images, \textit{bus} has $15,333$ annotations in RGB images, and $16,590$ annotations in infrared images, \textit{van} has $11,935$ annotations in RGB images, and $12,708$ annotations in infrared images, and \textit{freight car} has $13,400$ annotations in RGB images, and $17,173$ annotations in infrared image.

Since infrared images present higher contrast in low light conditions, they have more annotations than RGB images. According to different illumination levels, we divide the images in the DroneVehicle dataset into three scenarios: \textit{Day}, \textit{Night}, and \textit{Dark night}, which has $14,478$, $5,468$ and $8,493$ images respectively.
The ratio of the sum of \textit{Night} data and \textit{Dark night} data to \textit{Day} data is close to 1:1. As shown in the Fig.\ref{fig9}, in DroneVehicle, \textit{dark night} data mainly involve scenarios, such as parking lots, residential areas, roads without street lights; \textit{night} data mainly involve roads and blocks with lighting conditions.
In a dark scenario, many vehicles in the RGB image are difficult to be distinguished by human eyes, which are hard to be annotated precisely. But the same objects in an infrared image are much more distinct.
However, in some daytime scenarios, due to the lack of color information and texture details in infrared images, RGB images often have more complete annotations.

In drone-based practice, it is impossible for the camera to maintain a vertical downward viewing angle. Therefore, in addition to $15,475$ pairs of cross-modal images taken in vertical viewing angles, DroneVehicle also covers $12,964$ pairs of cross-modal images with three different angles ($15$°, $30$° and $45$°) and three different heights ($80m$,$100m$ and $120m$). We report the details with corresponding example images in different viewing angles and heights as shown in Fig.\ref{fig12}.

Each image in our DroneVehicle has an average of $16.76$ vehicles, of which the maximum number is $206$. In summary, this dataset covers different modalities, different scenarios, different lighting conditions, different viewing angles and heights, $etc.$, and contains a large number of finely annotated objects. This is the first and the largest dataset that can significantly promote the development of UAV-based RGB-Infrared vehicle detection.

\section{Method}
\label{sec:4}
In this work, we propose an uncertainty-aware cross-modality vehicle detection (UA-CMDet) framework. It contains an uncertainty-aware module (UAM) and a cross-modality detector (CMDet). Furthermore, we design an illumination-aware cross-modal non-maximum suppression strategy (IA-NMS) to better integrate the modal-specific information in the inference phase. The overall architecture of the proposed method is shown in Fig.\ref{fig2}.

\subsection{Uncertainty-Aware Module}
Uncertainty can be used to evaluate the credibility between different modalities~\cite{han2020trusted,geng2021uncertainty}.
Mount efforts have been developed to quantify the uncertainty by constructing a mathematical distribution~\cite{kendall2017uncertainties,sensoy2018evidential,van2020uncertainty}, which is quantified to evaluate indicators such as variance and standard deviation.
In this work, we propose an uncertainty-aware module (UAM) to quantify the cross-modal uncertainty in a task-driven manner. The left side of the Fig.\ref{fig2} shows the structure of UAM.

IoU (Intersection over Union) is used to evaluate the localization quality of the predicted box in object detection.
In the uncertainty-aware module, we calculate the IoU of the ground-truth bounding boxes in an RGB-infrared image pair and use it to quantify the localization difference of ground-truth bounding boxes in the two modalities.
Different from the IoU calculation in the horizontal bounding box, our cross-modal IoU ($CM\_IoU$) is performed within polygons.
The $B_{rgb}$ and $B_{infrared}$ represent the ground-truth bounding boxes of RGB and infrared images, respectively. The $area$ is a function for calculating the area of an arbitrary polygon. Finally, the $CM\_IoU$ can be calculated as,

\begin{align}\label{equ:IOU}
     CM\_IoU= \frac{area\left ( B_{rgb}\bigcap B_{infrared} \right )}{area\left ( B_{rgb}\bigcup B_{infrared} \right )}
\end{align}

In UAM, we first calculate the $CM\_IoU$ between the ground-truth bounding boxes of the RGB image and the Infrared image. Ideally, the $CM\_IoU$ should be close to 1. According to $CM\_IoU$, we can match the corresponding ground-truth boxes in two modalities, and then determine the index of the missing bounding boxes in the respective modality.

For the infrared modality, due to the lack of color information and texture details, the objects in the infrared image may have a confusing appearance, and it is easy to miss the annotations when manually annotating. In this case, we use the ground-truth boxes in the RGB modality to fill in these missing bounding boxes in the infrared modality.
Then, we assign an uncertainty weight $\omega_{inf}$ to each newly added bounding box in the infrared modality and set the weight of the original bounding box to 1.
Finally, the uncertainty weight $\omega_{I}$ of each object in the infrared modality can be calculated as,
\begin{align}
\omega_{I} =\begin{cases}
\omega_{inf} & \text{ if infrared object miss, } \\
1   & \text{ otherwise }
\end{cases}
\end{align}

Since objects with low visibility in the RGB image are missing annotations, it is difficult for certain bounding boxes in the infrared modality to find the corresponding position in the RGB modality.
In this case, these missing bounding boxes in the RGB modality are uncertain, so we assign an uncertainty weight $\omega_{rgb}$ to each of them. Meanwhile, we also use the ground-truth boxes in the infrared modality to fill in the missing bounding boxes in the RGB modality.
In addition, different lighting conditions also greatly affect the performance of the detector, especially for the RGB modality. For cross-modal object detector, different illumination levels also reflect the overall uncertainty of RGB images. Hence, we utilize the gray histogram to calculate the RGB illumination value.
In the night scenarios, we define the illumination value of the RGB image as the illumination uncertainty weight of the RGB modality, denoted as $\omega_{iv}$. In addition, it is difficult to ensure that the cross-modal image pairs are completely aligned pixel-by-pixel during image collection, which means $CM\_IoU$ is between $0$ and $1$. We take the object position in the infrared modality as a reference and set an alignment threshold as $\mu$. When $0<CM\_IoU<\mu$, we take the $CM\_IoU$ as the uncertainty weight of the misaligned object, denoted as $\omega_{cm\_iou}$, and assign it to the ground-truth bounding boxes of the object in the RGB modality.
Finally, the uncertainty weight $\omega_{R}$ of each object in the RGB modality can be calculated as,

\begin{align}
\omega_{R} =\begin{cases}
\omega_{rgb} & \text{ if RGB object miss, } \\
\omega_{cm\_iou}\times \omega_{iv} & \text{ if RGB object not aligned, }\\
\omega_{iv}   & \text{ if RGB object is aligned }
\end{cases}
\end{align}

\subsection{Uncertainty-Aware Cross-modality Detector}
Our uncertainty-aware cross-modality vehicle detector (UA-CMDet) includes a cross-modality detector (CMDet) and an uncertainty-aware module (UAM). Specifically, we choose RoITransformer~\cite{ding2019learning} as our basic oriented vehicle detector and modify it to a cross-modality detector (CMDet) to handle cross-modal inputs. Among them, CMDet is composed of RGB branch, infrared branch, and fusion branch, and we design a cross-modal fusion module to joint learning the fused feature and the respective knowledge of each modality. The uncertainty-aware module (UAM) provides the corresponding uncertainty weight for each detection head.

As shown in Fig.\ref{fig2}, the input of our UA-CMDet is a pair of RGB-Infrared images, which are feed to the feature extractors.
In this work, the ResNet~\cite{he2016deep} is taken as the feature extractor. We send the extracted feature maps of the two modalities into the cross-modal fusion module and obtain a set of feature maps with cross-modal knowledge. In the cross-modal fusion module, we first concatenate the feature maps of each modality in the channel dimension and then impose a $1\times1$ convolution layer to achieve dimensionality reduction and cross-channel information interaction.
Then the feature maps of the three branches are sent to their respective detection head. Among them, UAM provides the corresponding uncertainty weights for each detection head.
Considering the specific feature contained in the respective modality, we retrain the independent detection heads of the infrared modality and the RGB modality during training.
In the test phase, the outputs of the three detection heads are utilized to further enhance the detection performance.

We follow the detection head structure of \cite{ding2019learning} and design an Region Proposal Network (RPN)~\cite{ren2015faster} and an RoI Transformer module.
The RPN is responsible for proposing the horizontal proposals, and the RoI Transformer module is responsible for transforming the horizontal proposals into the oriented bounding boxes and performs fine classification and regression.
In each detection head, UAM can recalibrate the weights of the object bounding box regression involved in the RPN and RoI Transformer module, thereby reducing the regression loss of bounding boxes with uncertainty.
In UA-CMDet, the loss function $\mathcal{L}_{loc}$ of the object bounding box regression is as follows,

\begin{align}
\mathcal{L}_{loc}\left (t^{u},v,\omega\right)=\omega\sum_{i}^{}smooth_{L1}\left(t_{i}^{u}-v_{i}\right)
\end{align}

Where $i\in\left\{x,y,w,h,\theta\right\}$, $t^{u}$ represents the predicted result, and $u$ represents the ground-truth class. $v$ represents the ground-truth bounding-box regression target, $(x,y)$ denotes the center of the predicted result, and $(w,h)$ denotes the width and height of the predicted result. The $\theta$ gives the orientation of the predicted result. And $\omega$ represents the uncertainty weight, which can be calculated as,

\begin{align}
\omega =\begin{cases}
\omega_{R} & \text{ if it is the RGB branch, } \\
\omega_{I} & \text{ if it is the infrared branch, } \\
\omega_{F} & \text{ if it is the fusion branch }
\end{cases}
\end{align}

Among them, the uncertainty weight $\omega_{F}$ of each bounding box on the fusion branch is consistent with that on the infrared branch, which is calculated as,
\begin{align}
\omega_{F}=\omega_{I}
\end{align}

Taking the RGB branch as an example, We use a multi-task loss $\mathcal{L}_{rgb}$ to jointly train for classification and bounding-box regression:

\begin{align}
\mathcal{L}_{rgb}\left ( p,u,t^{u},v,\omega \right )=&\mathcal{L}_{cls}\left ( p,u \right )\notag
\\&+\lambda \left [ u\geq 1 \right ]\mathcal{L}_{loc}\left ( t^{u},v,\omega \right )
\end{align}

In which, $\mathcal{L}_{cls}$ uses the cross-entropy loss function, and $p$ represents the predicted probability of each class. The Iverson bracket indicator function $\left [ u\geq 1 \right ]$ evaluates to 1 when $u\geq 1$ and 0 otherwise. By convention the catch-all background class is labeled $u = 0$. The hyper-parameter $\lambda$ controls the balance between the two task losses. According to convention, all experiments use $\lambda = 1$. In UA-CMDet, the total loss function contains the loss functions of three branches. We perform a weighted sum on them:

\begin{align}
\mathcal{L}=\alpha \mathcal{L}_{rgb}+\beta \mathcal{L}_{inf}+\gamma \mathcal{L}_{fusion}
\end{align}

Where $\alpha,\beta,\gamma$ is the trade-off parameters. In all experiments, they are all set to 1, so that the respective modal weights are consistent.

\subsection{Illumination-Aware Cross-modal NMS}
Some information peculiar to a single modality may be lost in the cross-modal fusion phase, so it is generally difficult to achieve the best effect only using the detection results of the fusion branch. To avoid this problem, our UA-CMDet retains the independent detection head of the RGB modality and the infrared modality during the training phase and the inference phase. And in the inference phase, we can use the output of three independent branches to better integrate the modal-specific information.

In object detection, non-maximum suppression (NMS)~\cite{neubeck2006efficient} is often used to determine the final object bounding boxes. NMS sorts the candidate bounding boxes according to the classification probability output by the classifier. For a common object detector, the classification probability output by the classifier is often obtained through softmax, but softmax tends to inflate the probability of the predicted class~\cite{NEURIPS2018_a981f2b7}. This problem is more acute in cross-modal object detection. For example, in a dark scenario, the RGB branch can hardly accurately determine the true position of the object, so many false positive samples will be predicted. After softmax, these false positive samples will still get the corresponding classification probabilities.
When the prediction results of the three branches in the model are fused, the false positive samples predicted by the RGB branch will seriously affect the final fusion effect.

Considering that the RGB images is very sensitive to lighting conditions and NMS can remove redundant bounding boxes, we further propose a illumination-aware cross-modal non-maximum suppression strategy. In our illumination-aware cross-modal NMS (IA-NMS), $\mathcal{B}_{r} = \{b_{r1}, ..,b_{rN}\}$, $\mathcal{B}_{t} = \{b_{t1}, ..,b_{tN}\}$, $\mathcal{B}_{f} = \{b_{f1}, ..,b_{fN}\}$, $\mathcal{S}_{r} = \{s_{r1}, ..,s_{rN}\}$, $\mathcal{S}_{t} = \{s_{t1}, ..,s_{tN}\}$, $\mathcal{S}_{f} = \{s_{f1}, ..,s_{fN}\}$, among them, $\mathcal{B}_{r}$, $\mathcal{B}_{t}$ and $\mathcal{B}_{f}$ are the list of initial detection boxes for the RGB branch, the infrared branch and the fusion branch, respectively. And $\mathcal{S}_{r}$, $\mathcal{S}_{t}$ and $\mathcal{S}_{f}$ contain corresponding detection scores, respectively. Let $N_l$ be the NMS threshold. We use the illumination uncertainty weight $\omega_{iv}$ of the current RGB image to weight the detection scores of the candidate bounding boxes corresponding to the RGB modality, which can be expressed as $\mathcal{S}_{r}\leftarrow \mathcal{S}_{r} \times \omega_{iv}$. Then we merge all the candidate bounding boxes of the three branches together for NMS operation.

IA-NMS can reduce the interference of the RGB modal prediction results on the final detection results of the model in a dark scenario. As shown in Fig.\ref{fig2}, in the inference phase, UA-CMDet can realize the fusion of the results of the three detection head branches through IA-NMS. By convention, we set the NMS threshold $N_l$ to 0.1.

\begin{table}[t]
\caption{Statistics of the number of objects after the DroneVehicle dataset is split. R stands for RGB modality, I stands for infrared modality.}
    \label{tab:dataset split bounding boxes}
    \centering
	\renewcommand
	\arraystretch{1}
    \setlength{\tabcolsep}{1pt}
	\begin{tabular}{ccccccc}
        \toprule
		\multirow{2}{*}{Categories} &\multicolumn{2}{c}{Train}  &\multicolumn{2}{c}{Val}    &\multicolumn{2}{c}{Test} \\
		\cmidrule{2-7}
		& RGB & Infrared & RGB & Infrared & RGB & Infrared \\
        \midrule
		car &$246,700$  &$270,350$  &$18,965$  &$20,588$  &$124,114$ &$137,148$      \\
		truck  &$13,685$  &$15,833$  &$1,336$  &$1,470$  &$7,102$  &$8,657$    \\
		bus  &$10,421$  &$11,334$  &$751$  &$789$  &$4,161$  &$4,467$     \\
		van  &$7,275$  &$7,701$  &$700$  &$725$   &$3,960$  &$4,282$   \\
		freight car &$8,712$  &$11,193$   &$710$   &$918$  &$3,978$   &$5,062$   \\
		\bottomrule
	\end{tabular}
\end{table}

\section{Experiment}
\label{sec:5}
In this section, we present our experimental settings and extensive results with in-depth analysis. Firstly, we carry out ablation studies for the proposed method on the DroneVehicle dataset and then discuss the operations that affect the performance of the proposed method. Finally, the mAP of our method and the state-of-the-arts are reported, and the visualization of the detection results is discussed.

\subsection{Experimental Setting}

\begin{table*}
  \renewcommand\arraystretch{0.9}
      \caption{Ablation study on DroneVehicle dataset. Baseline is RoITransformer, UAM stands for uncertainty-aware module, R stands for RGB modality, I stands for infrared modality, the cross-modal fusion method that uses the element-wise-add operation is denoted as CM-E, and the cross-modal fusion method that uses the concatenate operation is denoted as CM-C. Where UA-CMDet$*$ represents that the UA-CMDet does not use IA-NMS and only uses the results of the fusion branch.}
      \begin{center}
      \begin{tabular}{C{1.9cm}C{1.2cm}C{0.9cm}C{0.9cm}C{0.8cm}C{1.3cm}C{0.8cm}C{1.49cm}C{0.8cm}C{0.8cm}C{0.8cm}C{1.0cm}}
      \toprule
        Methods  &Modality   &CM-E   &CM-C   &UAM    &IA-NMS    &car     &freight car     &truck        &bus     &van        &mAP \\
      \hline
      \specialrule{0em}{1pt}{1pt}
    	Baseline    &R    &$-$   &$-$  &$-$   &$-$     &$68.13$     &$29.08$     &$44.17$       &$70.55$     &$27.64$       &$47.91$   \\
    	Baseline    &R    &$-$   &$-$  &$\surd$  &$-$       &$68.11$     &$30.10$     &$43.91$       &$76.68$     &$28.02$  &$49.36$  \\
		Baseline    &I   &$-$   &$-$  &$-$  &$-$    &$88.85$    &$41.49$    &$51.53$     &$79.48$   &$34.39$       &$59.15$   \\
		Baseline    &I   &$-$   &$-$   &$\surd$   &$-$      &$89.24$    &$43.18$    &$51.29$     &$79.12$    &$34.78$    &$\mathbf{59.52}$  \\
      \hdashline\specialrule{0em}{ 1pt}{1pt}
		CMDet     &R+I   &$\surd$   &$-$  &$-$   &$-$       &$89.69$     &$46.60$     &$58.47$       &$80.63$     &$37.52$       &$62.58$  \\
		UA-CMDet$*$   &R+I   &$\surd$  &$-$   &$\surd$   &$-$   &$89.69$    &$47.28$    &$58.68$     &$80.76$     &$38.71$     &$63.02$  \\
		UA-CMDet   &R+I   &$\surd$   &$-$   &$\surd$   &$\surd$    &$87.97$    &$46.64$     &$57.71$   &$86.80$    &$38.26$    &$\mathbf{63.48}$  \\
      \hdashline\specialrule{0em}{1pt}{1pt}
		CMDet   &R+I   &$-$    &$\surd$   &$-$   &$-$     &$89.56$     &$47.99$     &$58.63$       &$80.75$     &$35.73$       &$62.53$  \\
		UA-CMDet$*$    &R+I   &$-$    &$\surd$   &$\surd$    &$-$     &$89.76$     &$47.15$     &$58.87$    &$80.65$     &$39.79$    &$63.25$  \\
		\textbf{Ours}    &R+I   &$-$   &$\surd$   &$\surd$   &$\surd$      &$87.51$    &$46.80$     &$60.70$    &$87.08$     &$37.95$    &$\mathbf{64.01}$ \\
      \toprule
      \end{tabular}
      \end{center}
      \label{tab:Ablation Study}
\end{table*}

\subsubsection{Implementation Details}
We utilize ResNet-FPN~\cite{lin2017feature}  as the backbone network, and the pretrained ResNet-50 model is used for initialization. Each image is randomly horizontally flipped with a probability of $0.5$ to increase the diversity. The whole network is optimized by SGD optimizer for $12$ epochs with a learning rate of $0.005$ and a batch size of $2$. Weight decay and momentum are set to $0.0001$ and $0.9$, respectively. We implement our codes with the Pytorch framework~\cite{NEURIPS2019_bdbca288} and conduct experiments on a workstation with two NVIDIA GTX1080Ti GPUs.
We have conducted experiments with different $\omega_{rgb}$, ranging from $0.01$ to $0.5$. Experimental results showed that the model achieved the highest mAP when the $\omega_{rgb}$ was set to $0.1$. For $\omega_{inf}$, we set it to $1$, because it was utilized as a reference model and only needed to compensate for the missing objects.
The alignment threshold $\mu$ measures the matching degree of the object positions between two modalities, we set $\mu$ to $0.8$ in our experiments.

\subsubsection{Baselines}
We compare our methods with 7 state-of-the-arts: RetinaNet(OBB)~\cite{lin2017focal}, Faster R-CNN(OBB)~\cite{Xia_2018_CVPR}, Faster R-CNN(Dpool)~\cite{dai2017deformable}, Mask R-CNN~\cite{he2017mask}, Cascade Mask R-CNN~\cite{cai2018cascade}, Hybrid Task Cascade$*$~\cite{chen2019hybrid}, RoITransformer~\cite{ding2019learning}. Specifically, for RetinaNet(OBB) and Faster R-CNN(OBB), OBB means that the detector has an oriented bounding box detection head. Further, we replace RoI Align in Faster R-CNN (OBB) with Deformable RoI Pooling, and call it Faster R-CNN (Dpool). For Mask R-CNN, Cascade Mask R-CNN, Hybrid Task Cascade$*$, they treat oriented object detection as a pixel-level classification problem. Where $*$ means that the semantic segmentation branch is removed from the Hybrid Task Cascade. For a fair comparison, all the detection results of our method and all baselines are obtained under identical experimental settings. We implement and evaluate all the algorithms in one unified code library modified from MMDetection~\cite{mmdetection}.

\subsubsection{Partition Protocol}
We split the dataset into the training set, validation set, and testing set. Among them, the training set contains $17,990$ RGB-Infrared image pairs, the validation set contains $1,469$ RGB-Infrared image pairs, and the remaining $8,980$ RGB-Infrared image pairs form the testing set. As shown in the Table. \ref{tab:dataset split bounding boxes}, in the training set, there are a total of $286,793$ vehicles in the RGB modality and $316,411$ vehicles in the infrared modality; in the validation set, there are a total of $22,462$ vehicles in the RGB modality and $24,490$ vehicles in the infrared modality; in the test set, there are a total of $143,315$ vehicles in the RGB modality and $159,616$ vehicles in the infrared modality. All experiments in this paper are trained on the training set and evaluated on the test set. It is worth noting that all training and validation sets in DroneVehicle will be open source.

\subsubsection{Evaluation Metric}
The standard metrics, Mean Average Precision (mAP) is adopted to evaluate the drone-based RGB-Infrared vehicle detection accuracy. The mAP measures the quality of bounding box predictions in the test set. Following~\cite{EveringhamThePascalVisual}, a prediction is considered as true positive if the IoU between the prediction and its nearest ground-truth annotation is larger than $0.5$.

\subsection{Ablation Study}
\subsubsection{Uncertainty-Aware Module}
To verify the effectiveness of UAM, we select RoITransformer as the baseline, directly add UAM to the baseline, and train the RGB modal detector and infrared modal detector respectively. As shown in Table. \ref{tab:Ablation Study}, UAM is effective to improve the mAP of the two modalities. Among them, the mAP of RGB modality is increased by $1.45\%$, and the mAP of infrared modality is increased by $0.37\%$. In addition, we used two feature fusion operations on the cross-modal detector (CMDet), namely the element-wise-add operation and the concatenate operation. For a fair comparison, UA-CMDet does not use IA-NMS and only uses the results of the fusion branch. And the CMDet uses the same training and inference configuration as UA-CMDet. When the element-wise-add operation is used, the mAP of the CMDet is $0.44\%$ lower than the UA-CMDet of the same fusion operation. When using the concatenate operation, the mAP of the CMDet is $0.72\%$ lower than the UA-CMDet. These experimental results prove that the uncertainty-aware module is effective.

\subsubsection{Cross-Modal Fusion}
To verify that the cross-modal fusion operation in UA-CMDet is effective, it can be seen from the Table. \ref{tab:Ablation Study} that even without IA-NMS, the mAP of our UA-CMDet is $13.89\%$ higher than the baseline with UAM in RGB modality, and $3.73\%$ higher than it in infrared modality. As a comparison, we used the element-wise-add operation to replace the cross-modal fusion operation in UA-CMDet and conducted experiments under the same conditions. The results show that its mAP is $0.23\%$ lower than the original UA-CMDet, but it is still better than the single-modality object detector.
In addition, for the cross-modal detector (CMDet), when the element-wise-add operation is used, the mAP of CMDet is $14.67\%$ higher than the baseline model trained in the RGB modality and $3.43\%$ higher than the baseline model trained in the infrared modality. When using the concatenate operation, the mAP of CMDet is $14.62\%$ higher than the baseline model trained in the RGB modality and $3.38\%$ higher than the baseline model trained in the infrared modality.
The above experiments prove that the cross-modal fusion operation is effective. According to the experiments in Table. \ref{tab:Ablation Study}, we also found that after using UAM and IA-NMS, the concatenate operation brings greater mAP gain to the model than the element-wise-add operation. Therefore, concatenate operation is selected as the cross-modal fusion method in UA-CMDet. After performing the concatenate operation on the features of the two modalities, we use a $1\times1$ convolution to ensure the consistency of the fusion features and the features of each single modality in the number of channels.

\subsubsection{Illumination-Aware NMS}
Under exactly the same experimental conditions, we conducted two sets of experiments to verify the effect of the illumination-aware NMS (IA-NMS) in UA-CMDet. The results are shown in the Table. \ref{tab:Ablation Study}. The first set of experiments uses concatenate operation (CM-C) and $1\times1$ convolution to achieve cross-modal feature fusion. When IA-NMS is not used, we directly use the output of the cross-modal fusion branch as the final detection results, and its mAP is lower than the model using IA-NMS by $0.76\%$. Then the second set of experiments directly used the element-wise-add operation (CM-E) to achieve cross-modal feature fusion. When IA-NMS is not used, we also use the output of the cross-modal fusion branch as the final detection results, and its mAP is $0.46\%$ lower than the model using IA-NMS. After using IA-NMS, in our UA-CMDet, the AP value of many categories has been improved. Moreover, no matter whether UA-CMDet uses CM-C or CM-E, the AP value of bus has increased by more than $6\%$. The above experiments prove that the illumination-aware NMS (IA-NMS) is effective.

\begin{table}
  \renewcommand\arraystretch{0.9}
      \caption{Detailed discussion in UA-CMDet. Baseline is RoITransformer, UAM stands for uncertainty-aware module, R denotes RGB modality, I denotes infrared modality, EWA is short for element-wise-add operation, and CAT is short for concatenate operation, SMO is short for \textit{Supplement of Missing Objects}, MA is short for \textit{Misalignment-Aware}, and IA is short for \textit{Illumination-Aware}. Where UA-CMDet$*$ represents that the UA-CMDet does not use IA-NMS and only uses the results of the fusion branch.}
      \begin{center}
      \begin{tabular}{C{2.9cm}C{1.1cm}C{0.6cm}C{0.45cm}C{0.45cm}C{0.85cm}}
      \toprule
        Methods  &Modality   &SMO  &MA  &IA  &mAP \\
      \hline
      \specialrule{0em}{1pt}{1pt}
		Baseline         &R        &$-$         &$-$        &$-$       &$47.91$  \\
		Baseline         &R        &$\surd$     &$-$        &$-$       &$48.33$  \\
		Baseline         &R        &$\surd$     &$\surd$    &$-$       &$48.94$  \\
		\bf Baseline+UAM    &R     &$\surd$     &$\surd$    &$\surd$   &$\mathbf{49.36}$  \\
      \hdashline\specialrule{0em}{1pt}{1pt}
		Baseline         &I         &$-$         &$-$        &$-$       &$59.15$  \\
		Baseline         &I     &$\surd$     &$-$        &$-$       &$59.42$  \\
		\bf Baseline+UAM    &I     &$\surd$     &$\surd$    &$-$   &$\mathbf{59.52}$  \\
      \hdashline\specialrule{0em}{1pt}{1pt}
		CMDet(EWA)      &R+I       &$-$     &$-$    &$-$   &$62.58$  \\
		CMDet(EWA)   &R+I       &$\surd$     &$\surd$    &$-$   &$62.82$  \\
		UA-CMDet$*$(EWA)   &R+I     &$\surd$     &$\surd$     &$\surd$   &$63.02$  \\
      \hdashline\specialrule{0em}{1pt}{1pt}
		CMDet(CAT)      &R+I       &$-$     &$-$    &$-$   &$62.53$  \\
		CMDet(CAT)   &R+I      &$\surd$     &$\surd$    &$-$   &$62.60$  \\
		\bf UA-CMDet$*$(CAT)   &R+I     &$\surd$     &$\surd$     &$\surd$   &$\mathbf{63.25}$  \\
      \toprule
      \end{tabular}
      \end{center}
      \label{tab:UAM Ablation Study}
\end{table}

\subsection{Disscussion}
To quantify the uncertainty weights of each modality, UA-CMDet mainly includes three operations: \textit{SMO} (Supplement of Missing Objects), \textit{MA}(Misalignment-Aware), and \textit{IA}(Illumination-Aware). To determine whether these operations are effective, we continue to explore them.

As shown in Table. \ref{tab:UAM Ablation Study}, in the RGB modality, the \textit{SMO} is added to the baseline, which includes assigning uncertainty weights to these missing objects. At this time, the mAP has increased by $0.42\%$ compared to the baseline. On this basis, after introducing the \textit{MA}, the mAP has increased by $0.61\%$. Finally, we add the \textit{IA} to the model and reach the highest mAP value. This mAP is $0.42\%$ higher than the mAP of the model without \textit{IA}. In the infrared modality, the \textit{SMO} is added to the baseline, which includes assigning uncertainty weights to these missing objects. At this time, the mAP has increased by $0.27\%$ compared to the baseline.
On this basis, after introducing the \textit{MA}, the mAP has increased by $0.10\%$, achieving the highest mAP in a single modality.
For the cross-modal detector (CMDet), we conducted experiments on two cross-modal fusion operations.
For the element-wise-add operation, after the \textit{SMO} and the \textit{MA} are added to the CMDet(EWA), the mAP of the model is increased by $0.24\%$. And after added the \textit{IA}, the mAP of the model has been further increased by $0.20\%$.
For the concatenate operation, after the \textit{SMO} and the \textit{MA} are added to the CMDet(CAT), the mAP of the model is increased by $0.07\%$. And after added the \textit{IA}, the mAP of the model has been further increased by $0.65\%$.
This all proves that the three operations we proposed are effective and can be embedded in any RGB-Infrared cross-modality vehicle detection framework.

\begin{table*}
  \renewcommand\arraystretch{0.9}
      \caption{Evaluation on our DroneVehicle Dataset. Where $*$ means that the semantic segmentation branch is removed from the Hybrid Task Cascade.}
      \begin{center}
      \begin{tabular}{C{4.2cm}C{2.4cm}C{1.2cm}C{1.48cm}C{1.2cm}C{1.2cm}C{1.2cm}C{1.2cm}}
      \toprule
        Methods  &Modality   &car     &freight car     &truck        &bus     &van        &mAP \\
      \hline
      \specialrule{0em}{1pt}{1pt}
		RetinaNet(OBB)~\cite{lin2017focal}      &RGB      &$67.50$      &$13.72$      &$28.24$     &$62.05$     &$19.26$     &$38.16$  \\
		Faster R-CNN(OBB)~\cite{Xia_2018_CVPR}      &RGB       &$67.88$      &$26.31$      &$38.59$     &$66.98$     &$23.20$     &$44.59$  \\
		Faster R-CNN(Dpool)~\cite{dai2017deformable}   &RGB     &$68.23$   &$26.40$      &$38.73$       &$69.08$     &$26.38$       &$45.76$  \\
		Mask R-CNN~\cite{he2017mask}    &RGB      &$68.52$     &$26.83$   &$39.84$       &$66.75$     &$25.35$       &$45.46$  \\
		Cascade Mask R-CNN~\cite{cai2018cascade}         &RGB      &$68.00$    &$27.25$      &$44.67$       &$69.34$     &$29.80$       &$47.81$  \\
		Hybrid Task Cascade$*$~\cite{chen2019hybrid}    &RGB      &$67.89$      &$27.22$       &$44.55$     &$70.22$    &$28.61$     &$47.70$  \\
		RoITransformer~\cite{ding2019learning}             &RGB        &$68.13$     &$29.08$     &$44.17$       &$70.55$     &$27.64$       &$47.91$  \\
      \hdashline\specialrule{0em}{1pt}{1pt}
		RetinaNet(OBB)~\cite{lin2017focal}      &Infrared       &$79.86$      &$28.05$      &$32.84$     &$67.32$     &$16.44$     &$44.90$  \\
		Faster R-CNN(OBB)~\cite{Xia_2018_CVPR}         &Infrared      &$88.63$      &$35.16$     &$42.51$    &$77.92$    &$28.52$       &$54.55$  \\
		Faster R-CNN(Dpool)~\cite{dai2017deformable} &Infrared       &$\mathbf{88.94}$   &$36.79$      &$47.91$       &$78.28$     &$32.79$       &$56.94$  \\
		Mask R-CNN~\cite{he2017mask}  &Infrared      &$88.77$  &$36.63$      &$48.86$       &$78.38$     &$32.16$       &$56.96$  \\
		Cascade Mask R-CNN~\cite{cai2018cascade}       &Infrared      &$81.00$    &$38.97$      &$47.18$       &$79.32$     &$33.00$       &$55.89$  \\
		Hybrid Task Cascade$*$~\cite{chen2019hybrid}      &Infrared       &$88.57$    &$42.85$      &$47.71$       &$79.46$     &$34.16$       &$58.55$  \\
		RoITransformer~\cite{ding2019learning}           &Infrared         &$88.85$    &$41.49$    &$51.53$     &$79.48$   &$34.39$       &$59.15$  \\
      \hdashline\specialrule{0em}{1pt}{1pt}
		\bf UA-CMDet(Ours)      &RGB+Infrared    &$87.51$    &$\mathbf{46.80}$     &$\mathbf{60.70}$    &$\mathbf{87.08}$     &$\mathbf{37.95}$    &$\mathbf{64.01}$  \\
      \toprule
      \end{tabular}
      \end{center}
      \label{tab:Evaluation on our DroneVehicle Dataset}
\end{table*}

\begin{figure*}
    \centering
    \subfigure[RGB Modality.] {
        \label{fig:a}
        \includegraphics[width=0.98\columnwidth]{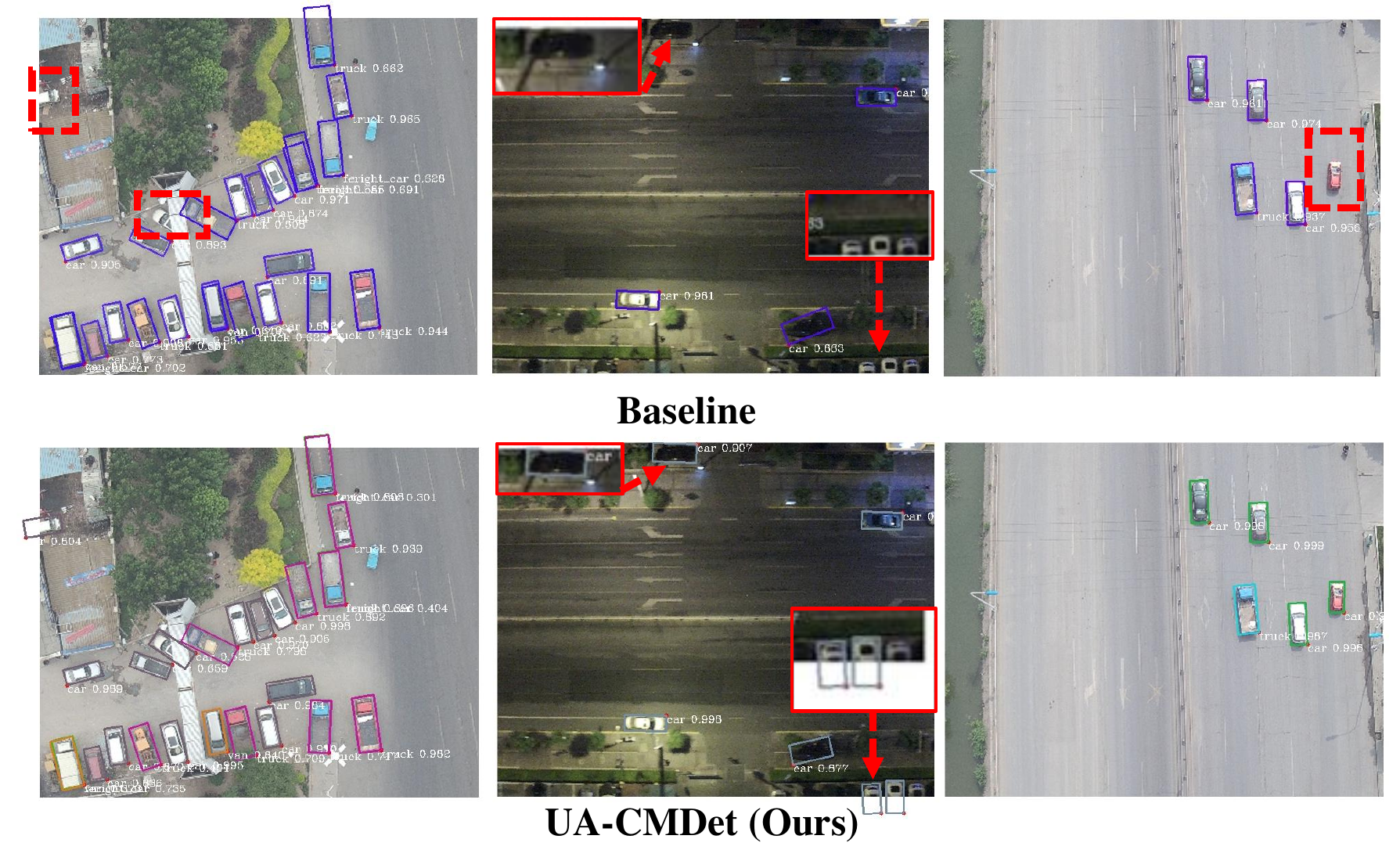}
    }
    \subfigure[Infrared Modality.] {
        \label{fig:b}
        \includegraphics[width=0.98\columnwidth]{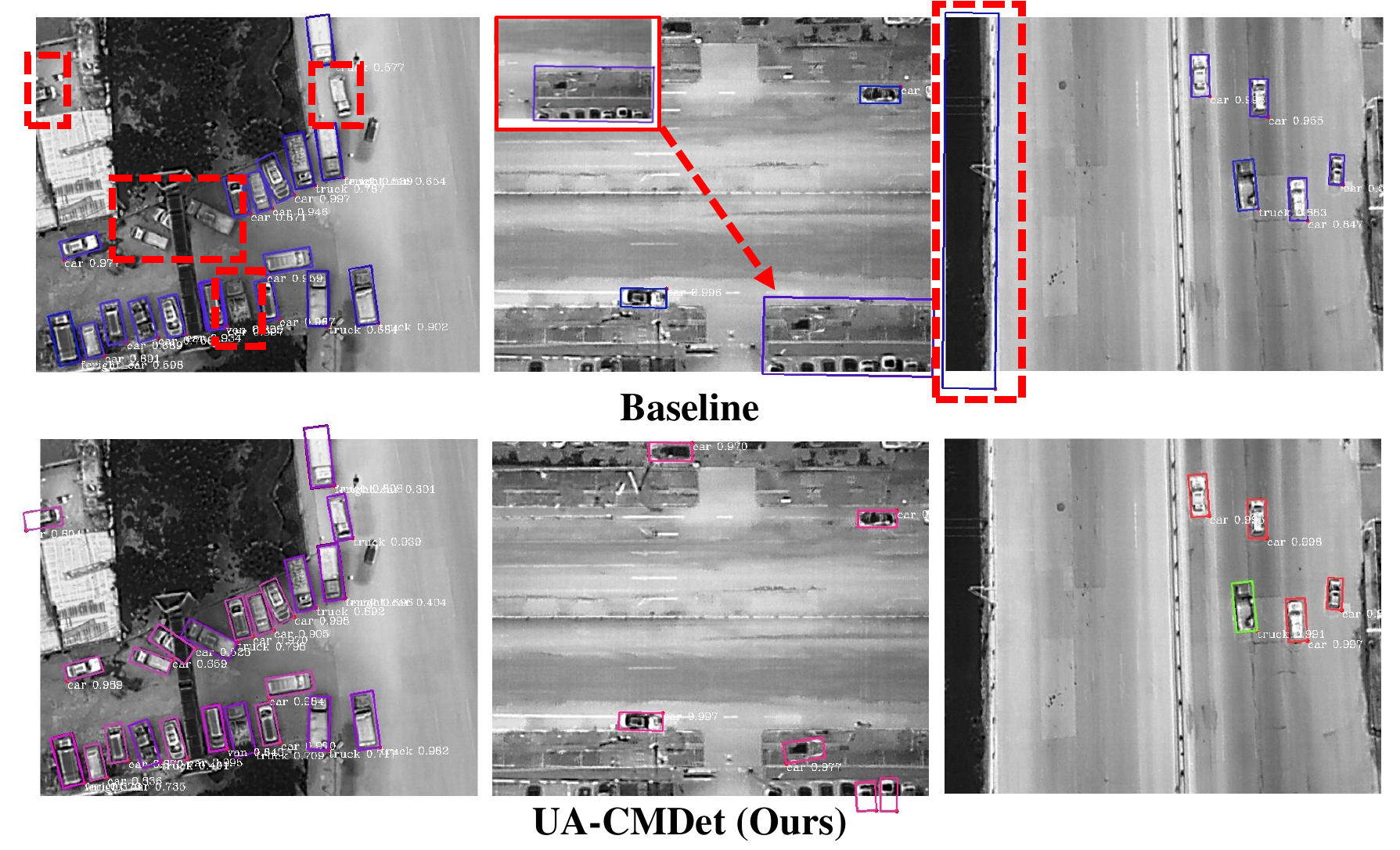}
    }
    \caption{Qualitative analysis of UA-CMDet on DroneVehicle. (a) and (b) are the qualitative analysis in the RGB modality and infrared modality respectively. The first row of (a) shows the detection results of the baseline in the RGB modality, and the second row shows the detection results of UA-CMDet in the RGB modality. The first row of (b) shows the detection results of the baseline in the infrared modality, and the second row shows the detection results of UA-CMDet in the infrared modality. The area selected by the red dashed box in the figure represents the error detection and missed detection. The solid red box represents the magnification effect of the area of interest in the image.}
    \label{fig15}
\end{figure*}

\subsection{Comparisons}
\subsubsection{Quantitative evaluation}
We compare our method with state-of-the-art methods on the DroneVehicle dataset under the same settings. The results are shown in the Table.~\ref{tab:Evaluation on our DroneVehicle Dataset}.
Since existing methods~\cite{lin2017focal,Xia_2018_CVPR,dai2017deformable,he2017mask,cai2018cascade,chen2019hybrid,ding2019learning} are designed for single modality, we train them on single modality (RGB or infrared modality). RoITransformer achieved the advanced mAP among these single-modality object detectors. Compared with them, our UA-CMDet achieves superior performance with the highest mAP. And compared with the RoITransformer, the mAP of our UA-CMDet has increased by $16.10\%$ and $4.86\%$ in RGB modality and infrared modality, respectively.

Compared to the highest accuracy of each category in the RGB modality, our method improves the accuracy of four categories by more than $16\%$. And compared to the highest accuracy of each category in the infrared modality, our method improves the accuracy of all categories by more than $3.5\%$.
Specifically, in the RGB modality, our method improves the accuracy of car by $18.99\%$, the accuracy of freight car by $17.72\%$, the accuracy of truck by $16.03\%$, and the accuracy of bus by $16.53\%$. But our method only improves the accuracy of van by $8.15\%$. The possible reason is that the appearance similarity between van and car is higher, which makes it difficult for the object detector to distinguish them. And in the infrared modality, our method improves the accuracy of freight car by $3.95\%$, the accuracy of truck by $9.17\%$, the accuracy of bus by $7.6\%$, and the accuracy of van by $3.56\%$. However, we find that the accuracy of car decrease. Here we give an explanation. The predicted bounding boxes generated by the three branches of UA-CMDet will be sent to the NMS together. Since there is more overlap between each other, some positive sample objects may be filtered out by NMS.

\begin{figure*}[!t]
	\centering	
	\includegraphics[width=2\columnwidth]{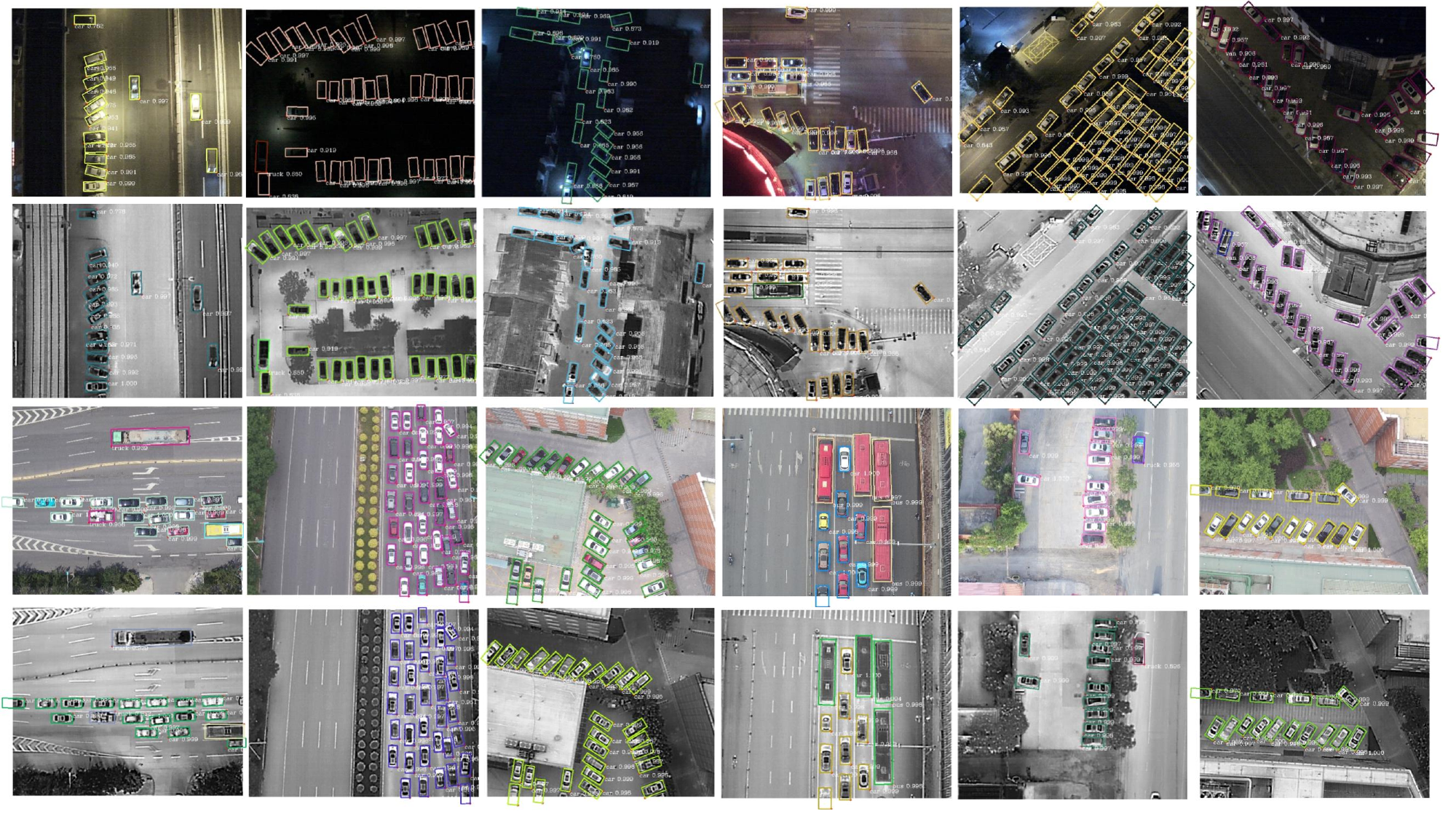}
	\caption{Visualization of UA-CMDet detection results on DroneVehicle. The first row shows the detection results in the night scenarios. The third row shows the detection results in the daytime scenarios. The second row and the fourth row respectively represent the detection results of the corresponding infrared images.}
	\label{fig8}
\end{figure*}

\subsubsection{Qualitative evaluation}
We use the RoITransformer as the baseline, and then qualitatively evaluate our UA-CMDet on the DroneVehicle dataset. In Fig.~\ref{fig1}~\subref{fig:c}, we have shown that UA-CMDet can cope with the problem of uncertain vehicle locations in RGB images under poor lighting conditions. And Fig.~\ref{fig1}~\subref{fig:d} also shows the correct detection results of UA-CMDet for the ``ghost shadows'' and confusing rectangle objects in infrared images. Next, we continue to conduct qualitative evaluations on more examples.
In Fig.~\ref{fig15}~\subref{fig:a}, the first row shows the detection results of the baseline in the RGB modality, and the second row shows the detection results of UA-CMDet in the RGB modality. In Fig.~\ref{fig15}~\subref{fig:b}, the first row shows the detection results of the baseline in the infrared modality, and the second row shows the detection results of UA-CMDet in the infrared modality. The area selected by the red dashed box in the figure represents the error detection and missed detection. The solid red box represents the magnification effect of the area of interest in the image.
But the results of the second and fourth rows show that our UA-CMDet is not bothered by these problems. UA-CMDet can accurately detect each object in the two modalities, which proves that our method can effectively overcome the problems existing in a single modality and improve the accuracy of vehicle detection. The main reason for this success is that our UA-CMDet can effectively learn the information of the two modalities, and use this information to make up for the shortcomings of the single-modality.

Fig.\ref{fig8} shows the visualization of our UA-CMDet detection results on the DroneVehicle dataset. Among them, the first row and the third row show the visualization of the detection results on the RGB image. The first row mainly shows the detection results in the night scenarios, and the third row mainly shows the detection results in the daytime scenarios. The second and fourth rows show the visualization of the detection results on the corresponding infrared image. It can be seen from Fig.\ref{fig8} that our detection results are very encouraging, which also proves the effectiveness of our method.

\section{Conclusion}
\label{sec:6}
In this paper, we constructed a large-scale drone-based RGB-Infrared vehicle detection dataset (DroneVehicle), which is the first and the largest cross-modal dataset that makes vehicle detection possible in complex aerial scenes. We sincerely hope that DroneVehicle can contribute to the computer vision community.
Considering the great gap between RGB and infrared images, we proposed an uncertainty-aware cross-modality vehicle detection (UA-CMDet) framework, which joints the uncertainty information of two different modalities to better extract cross-modal effective information. Among them, an uncertainty-aware module (UAM) was designed to quantify the uncertainty weights of RGB modality and infrared modality. We further proposed an illumination-aware NMS (IA-NMS) to integrate the modal-specific information in the inference phase.
Extensive experiments have validated the effectiveness of the proposed framework and its internal modules. Our proposed framework achieves superior performance against state-of-the-arts on DroneVehicle.

In this work, we mainly focused on uncertainty quantification of RGB modality and infrared modality in a unified cross-modal fusion framework. Since our DroneVehicle was collected from the real world, the long tail problem still exists in the dataset, which will also affect the performance of the vehicle detector.
In the future, we will take the long tail data distribution into consideration, and explore a more effective framework to further increase the detection accuracy and improve the robustness on tail objects.
More importantly, we sincerely hope that more researchers can contribute to this field and use DroneVehicle to promote the development of drone-based cross-modality vehicle detection for smart city traffic management.


\ifCLASSOPTIONcaptionsoff
  \newpage
\fi




\normalem

\bibliographystyle{IEEEtran}
\bibliography{TITS}

%





\end{document}